\documentclass{article}

 \PassOptionsToPackage{numbers, compress}{natbib}

 \usepackage[preprint]{neurips_2020}

\usepackage[utf8]{inputenc} 
\usepackage[T1]{fontenc}    
\usepackage{hyperref}       
\usepackage{url}            
\usepackage{booktabs}       
\usepackage{amsfonts}       
\usepackage{nicefrac}       
\usepackage{microtype}      

\usepackage{graphicx}

\usepackage{subcaption}

\title{ Modular network for high accuracy object detection }

\author{Erez Yahalomi}

\begin{document}

\maketitle




\pagenumbering{arabic}

\begin{abstract}

We present a novel modular object detection convolutional neural network that significantly improves the accuracy of object detection.  The network consists of two stages in a hierarchical structure. The first stage is a single network that detects general classes.  The second stage consists of several separate networks to refine the classification and localization of each of the general class objects. Compared to a state-of-the-art object detection network, the classification error in the modular network is improved by approximately 3-5 times, 12\% to 2.5 \%-4.5\%. This network is easy to implement and has a 0.94 mAP.  The network architecture is a platform to improve the accuracy of many detection networks and other types of deep learning networks. We show experimentally and theoretically that a deep learning network that is initialized by transfer learning become more accurate as the number of classes it later trained to detect become smaller.

\end{abstract}

\section{Introduction}

There is constant effort to increase the accuracy of deep learning networks for object detection. A major topic in object detection is fine-grained \citep{Krause_2015_CVPR, Hariharan_2015_CVPR, Singh_2016_CVPR, DBLP:journals/corr/HowardZCKWWAA17}   detection for distinguishing differences between similar object classes. In this paper, we present a novel, highly accurate deep learning network for computer vision object detection, in particular, for fine-grained object detection. 
Our contribution in this paper is a new object detection modular network of two stages in hierarchical structure, from detection of general classes to more detailed classes. The first stage is one deep learning object detection network for detecting multi-class objects where the classes are general. The second stage consists of several separate deep learning object detection networks, each trained to detect only fine-grained classes that belong to one of the general classes of the first stage network. Images belonging to one of the general classes detected in the first stage  are passed on to the appropriate network in the second stage for more detailed identification of an object's type and location. We compared the results of our modular object detection network to a state-of-the-art object detection network, which was trained to detect the same classes as the modular network. The experiments showed that the modular network has significantly higher accuracy.
 We show both experimentally and theoretically that a deep learning network designed to detect a smaller number of classes and initially trained by transfer learning is more accurate than a network trained to detect more classes.
The modular network architecture suggested in this paper can be used to increase the accuracy of state-of-the-art object detection networks by integrating them as the building blocks of this modular network without changing the intensive optimizations carried out on their architecture.  Other types of networks can improve their accuracy, as well, by being inserted as building blocks into this modular network platform.

\section{Related Work}
We are the first to propose a modular network with a hierarchical structure for fine-grained object detection whose consist entirely on deep learning object detection networks \citep{rplan}. An additional novelty of our modular network is that input images are passed on for detection to the appropriate second stage networks based on the objects classes and their confidence score detected by the first stage object detection network.

\subsection{Object detection}
Notable convolutional neural networks for object detection are  \citep{journals/corr/Girshick15,journals/corr/LiuAESR15,DBLP:journals/corr/RedmonDGF15,43022} and Faster R-CNN \citep{DBLP:journals/corr/RenHG015} which, consists of a classification network, a region proposal network which divides the image into rectangular regions, followed by regression for additional accuracy in classification and location. Most of the state-of-the-art object detection networks include a core image classification network, such as Alexnet \citep{Krizhevsky2012ImageNetCW}, VGG \citep{simonyan2014very} or Resnet \citep{DBLP:journals/corr/HeZRS15}. These networks use transfer learning based on the training on a large image data, set such as Imagenet  \citep{10.1007/s11263-015-0816-y, Deng2009ImageNetAL}  and Coco \citep{lin2014microsoft}. 
\subsection{Hierarchical structures} 
Hierarchical structures appear in many forms in computer vision \citep{DBLP:journals/nn/Fukushima88,5995720}. Jarrett et al. \citep{inproceed} present a hierarchical feature extraction and classification system with fast (feed-forward) processing. The hierarchy stacks one or several feature extraction stages, each of which consists of filter bank layer, non-linear transformation layers, and a pooling layer . Salakhutdinov et al \citep{5995720} presented a hierarchical classification model that allows rare objects to borrow statistical strength from related objects that may have many training instances. They use hierarchical classification model where parameters of each class are given by the sum of parameters along the tree.

\section{The modular network}

\subsection{Modular network architecture}
We present in this paper a new modular and hierarchical object detection network. The network consists of two stages. The first is a deep learning, object detection network trained to detect predetermined general classes, and the second stage consists of several, deep learning, object detection networks, each trained independently on more fine-grained classes belonging to the same single general class of the first stage network. All the building block networks inside the modular network are trained on negative images as well.  All the object detection networks inside the modular networks are whole and separate object detection networks.

Each deep learning network in the modular network independently goes through the complete object detection process of training and inference.

The full image data set for inference is inserted into the first stage network. If an object in an image is detected as belonging to one of the network general classes, the initial image without changes is passed onto inference by a second stage network trained to detect sub-classes of this class. The purpose of the second stage network is to distinguish between objects of similar classes making more detailed classification and more accurate locations of each object in the image. Each sub-network inside the modular network is initialized by transfer learning weights \citep{DBLP:journals/corr/HuhAE16,Karpathy_2014_CVPR,oquab2014learning,IEEE_Transactions,DBLP:journals/corr/YosinskiCBL14} trained on ImageNet database. Figure.1 shows the modular network of our experiment. The building blocks or sub-networks of the modular network are Faster-RCNN (frcnn for short) networks \citep{DBLP:journals/corr/RenHG015}. In the first stage, there is a single network trained to detect five general objects classes. Based on the first stage network output images are passed onto fine-grained detection in the second stage at the appropriate network trained to detect the detailed classes belonging to the general classes determined by the first stage. 

\begin{figure}[h!]
\centering
\includegraphics[width=0.65\textwidth,height=0.25\textheight]{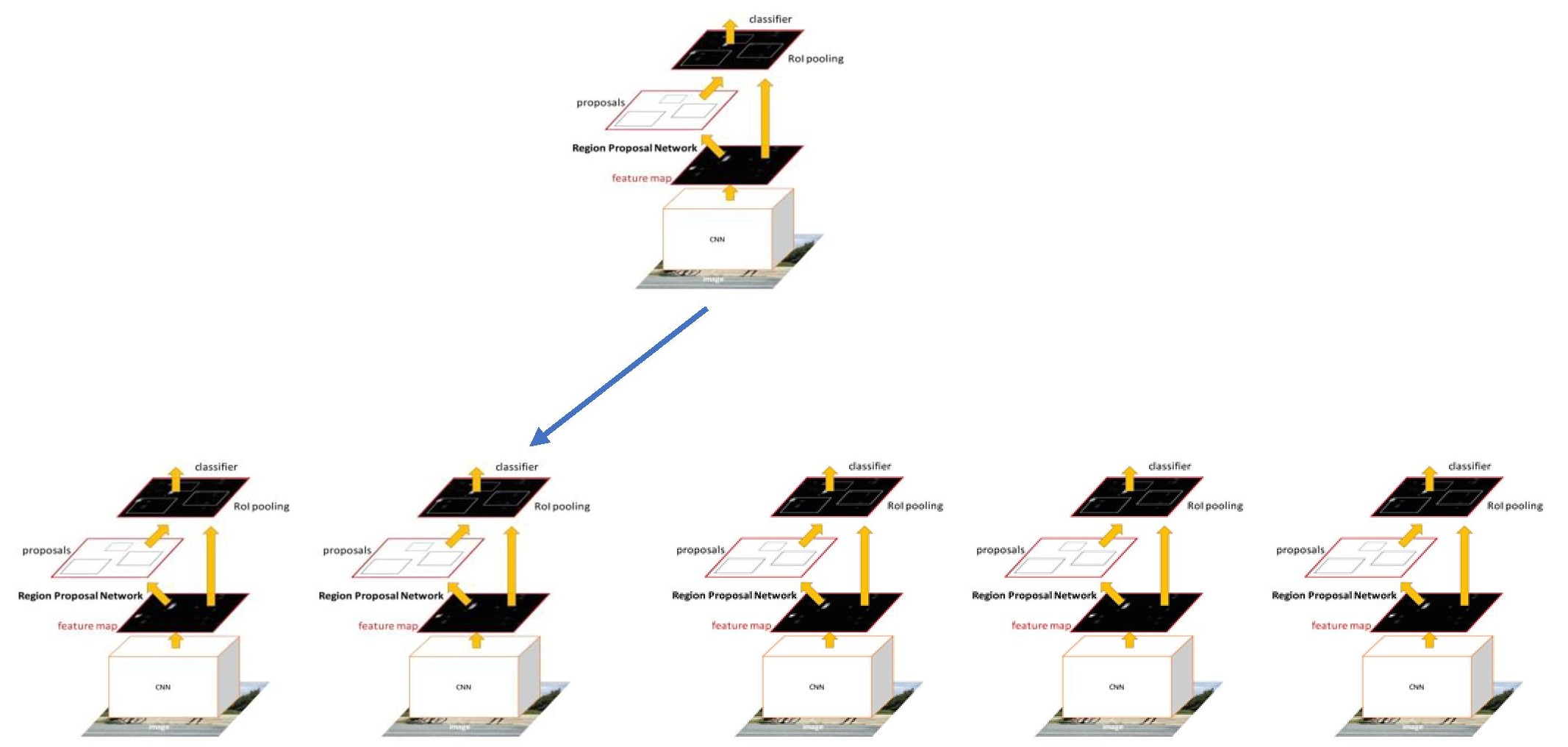} 
\label{graph3} 
\caption{A modular network whose first stage is a single deep learning network frcnn trained to detect five general classes. Its second stage consists of five separate frcnn, each trained to detect two distinct sub-classes of one of the general classes.}
\end{figure}

One of the main reasons for better accuracy over a regular multi-class network is that each of the networks inside this modular network is designated to detect fewer classes than a regular multi-class network. 

For a very large number of classes, a possible further modification of the modular network would be to add additional hierarchical stages.

\subsection{Algorithm and the modular network construction}

To detect multiple classes, we use on the first stage an object detection network trained by transfer learning. Similar classes are merged into a general class. The first stage network, is trained to detect general classes $C_i$ and additional negative images that do not belong to any of these general classes. For each of the general classes, we train a second stage network on the same training images of this general class and on negative images. This time, we sort and label the training images with fine-grained classes all belonging to this general class. 
Images are then input into the first stage network for inference. Input images with object detected as belonging to a general class on the basis of the network 's confidence score of the detected object are passed onto the second stage network dedicated to that class for fine-grained object classification and location.

\subsection{Advantages and risk of the modular network}
The advantages of the modular network are:
In each of the sub-convolutional neural networks inside the modular network, there are fewer classes than in a regular network designated to detect the same number of classes as the whole modular network. Thus, there are more features, filters, and network parameters dedicated to the detection of each class, resulting in better accuracy.  A small number of features for each class allows less distinction in detection of similar classes as well as errors in detecting rare class objects. When the number of features is small, features are formed to identify multiple classes, adding errors in fine-grained object detection.
Fewer classes in the object detection network means potentially fewer bounding boxes of detected objects in the image, which gives fewer errors in identifying the objects and finding their locations. The advantage of the hierarchical structure of the modular network compared to detection by many  networks  each detect  few classes with no connection  between the networks is that the hierarchical structure of the modular network drastically cuts down the number of required inferences, as they are arranged in a tree structure. A risk of the modular network is: Assuming we use the same type of object detection network as the multi class network as the building block network of the modular networks. If the multi-class network has low accuracy, then it is preferred since the building blocks networks inside the modular network should have a very large improvement in accuracy compared to the multi-class network to make the modular network work  more accurate. The condition the accuracy of the modular network will be better than a multi class network is,when: \begin{equation}
a < (a+\Delta_{1})(a+\Delta_{2})
\end{equation}a - represents the multi-class network accuracy;  $\Delta_{1}$ is- the improvement in accuracy of the first stage of the modular network compared to the multi-class network; and  $\Delta_2$ shows  the improvement in accuracy of the second stage compared to the multi-class network.
 Most state-of-the-art object detection networks are accurate enough to use them as the building block network for the modular network allowing a modular network with higher accuracy compared to the selected state-of-the-art object detection network. The risk of the modular network is the detection of false negatives in the first stage network. This may reduce accuracy, as some images with true objects may be omitted from the input of the second stage network.  To deal with this problem, we designed a second version of the modular network specifically for  images sequence \citep{inbook, articl,qiu2019visual} where the same object is assumed to appear in more than one image. The network architecture of this version, denoted as Modular Network v.2  is the same as Modular Network v.1,  the difference is that once an object of general class is detected in the first stage all of the inference images set is sent for inference to the appropriate fine-grained network in the second stage. In this way, the loss of accuracy due to false negative detection in the first stage is reduced.\\ \\

\section{Convolutional neural network classification error model.}

This model describes how reducing the number of classes for detection in a convolutional neural network (CNN) reduces the network classification error. Each of the building block networks inside the modular network has fewer classes than the regular multi-class network. Let x= \{ $x_1$…$x_f$\}  be the features space. Let c be a set of classes c=\{$c_0$…$c_n$\}. Every detection of an object in an image is defined by a set of  features that are active if this object appears in the image. For example, the features set \{$x_m$…$x_p$\} identifies objects belonging to class $C_1$. N represents the total number of features of the designated classes that the CNN can identify. L and T are the numbers of features of the designated classes that the CNN can identify based on transfer learning and fine tuning \citep{DBLP:journals/corr/YosinskiCBL14,inproceed,5995720}, respectively, where each feature belongs to a single class.  U is the number of features that the CNN can identify that are common to several classes. N= L+T+U. When each of the designated classes has a similar number of training images, $S$ - the number of features detecting a designated class, is approximated as $S\approx\frac{{L+T}}{n}+U$.  In this approximation, the number of features for detecting a single designated class is inversely related to n, the number of the CNN designated classes. The smaller the n, the more features there are for detecting the  designated class, making this  class object  detection more accurate. The parameters that determine \textit{ sup} K - the upper bound of features of all the classes that a CNN can identify include: r - the number of parameters in the CNN; a – the number of filters; d – the size of the filters; h – the number of filter channels; and q – the number of layers in the CNN. These parameters are constant for each network. In this model, every CNN has an upper bound of the total number of features, \textit{ sup} K(r,a,d,h,q), that it can identify without increasing the classification errors. Classification error caused by having more features than the upper bound number can be, for example, from two channels in the same filter, where the weight patterns formed in each channel detect different  classes. Suppose each of these channels doing a convolution with its  respective feature map. Where the different objects classes features on the different feature  map are located in similar locations in the input feature maps.            The two output feature maps of the two channel  patterns can have partial overlap in their  locations. 

Let A and B be matrices presented the two channels output features maps. Some of the feature weights in Matrix A can have the same pixel coordinates as the weights of the feature in  Matrix B.
   \begin{equation}
\sum{_{i,j\in G} ({|A|}_{i,j}+{|B|}_{i,j}})>\sum{_{i,j\in G}}{|A|}_{i,j} 
\end{equation}

In eq.2,  i and j are the raw and column coordinates of the elements in matrices A or B.  G is a set of all the coordinates that are active in both A and B matrices. Eq.2 indicates that when adding the elements of these coordinates from both feature maps the sum is no longer presenting a feature map of the object detected in matrix A but a deformation caused by the features sum of two different classes objects. This can cause classification error. 

From these it is obvious that increasing the number of filters a in deep learning network layers increase the network accuracy or able to identify more classes without reducing the accuracy. Since it will able to spread the different features kernels on more filters.

To estimate the classification error Bayes error is used \citep{article1,article2,568732,biobayes}. As an example, we analyzed the classification of two fine-grained classes, $C_1$ and $C_0$. According to Bayes error estimation, there is a probability that feature $x_i$ appears in the feature map when there is an object of class $C_0$ in the image. There is also a probability density that feature  $x_i$ is activated when an object of class $C_1$ is in the image. The classification error caused by feature $x_i$ is the smallest probability density between these two probabilities densities. The sum of the smallest probability densities for all the features that activated by the two classes object is the classification error. Assuming the probability densities to be activated by objects of classes $C_1$ or $C_0$ are known for each of the features in the network, the probability for classification error is described in Equation 2, where P($C_0$) and P($C_1$) are the  prior probability densities of classes $C_0$ and $C_1$, respectively. P($x_i|$ $C_0$) and P($x_i|$ $C_1$) are the conditional probability densities that feature $x_i$ is active given the class is $C_0$ or $C_1$, respectively. An additional criterion in Equation 2 is the significance of the feature $x_i$ in the classification. Because if an active feature does not influence the classification of an object, it does not contribute to the classification probability of the object class. Feature $x_i$ weights for classes $C_0$ and $C_1$ are denoted by $w_i(C_0)$ and $w_i(C_1$), respectively. The values of the weights $w_i(C_0)$ and $w_i(C_1$ ) are based on how many times feature $x_i$ is essential for classification out of all the times this feature was activated by the class objects.
  \begin{equation}
P_{error}=\sum_{i=1}^{N_f}min(P(x_i|C_0)P(C_0)w_i(C_0), P(x_i|C_1)P(C_1)w_i(C_1))\end{equation}
The probability densities of the features are presented in discrete values, which we approximate as a continues graph.The Graphs in figure.2a,2b, present the features probabilities densities to be activated by objects of classes $C_0$ and $C_1$. The X-axis is the feature range, i - is the  filter index number. The Y-axis presents the probability density that a feature is activated. In the graph, all features with the probability of matching a particular class are in the same area on the X-axis. Features with a probability of matching the two classes are displayed in the graph in a shared area for both classes. The Bayes classification error is the sum, or integration, of the minimal probability densities of every feature within the mutual area, which is the overlapping of the classes $C_0$ and $C_1$ curves.

\begin{figure}[h!]
\centering
\begin{subfigure}{.52\textwidth}
  \centering
  \includegraphics[width=1.3\textwidth, height=0.27\textheight]{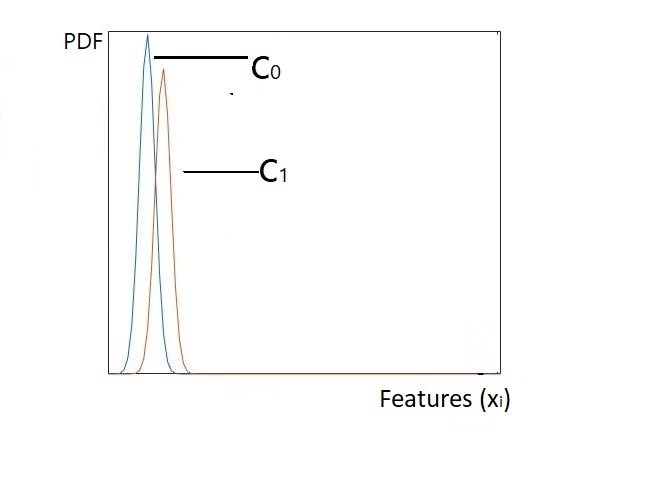}
  \caption{Ten classes network}
  \label{fig:sub1}
\end{subfigure}%
\begin{subfigure}{.5\textwidth}
  \centering
  \includegraphics[width=1.0\textwidth, height=0.27\textheight]{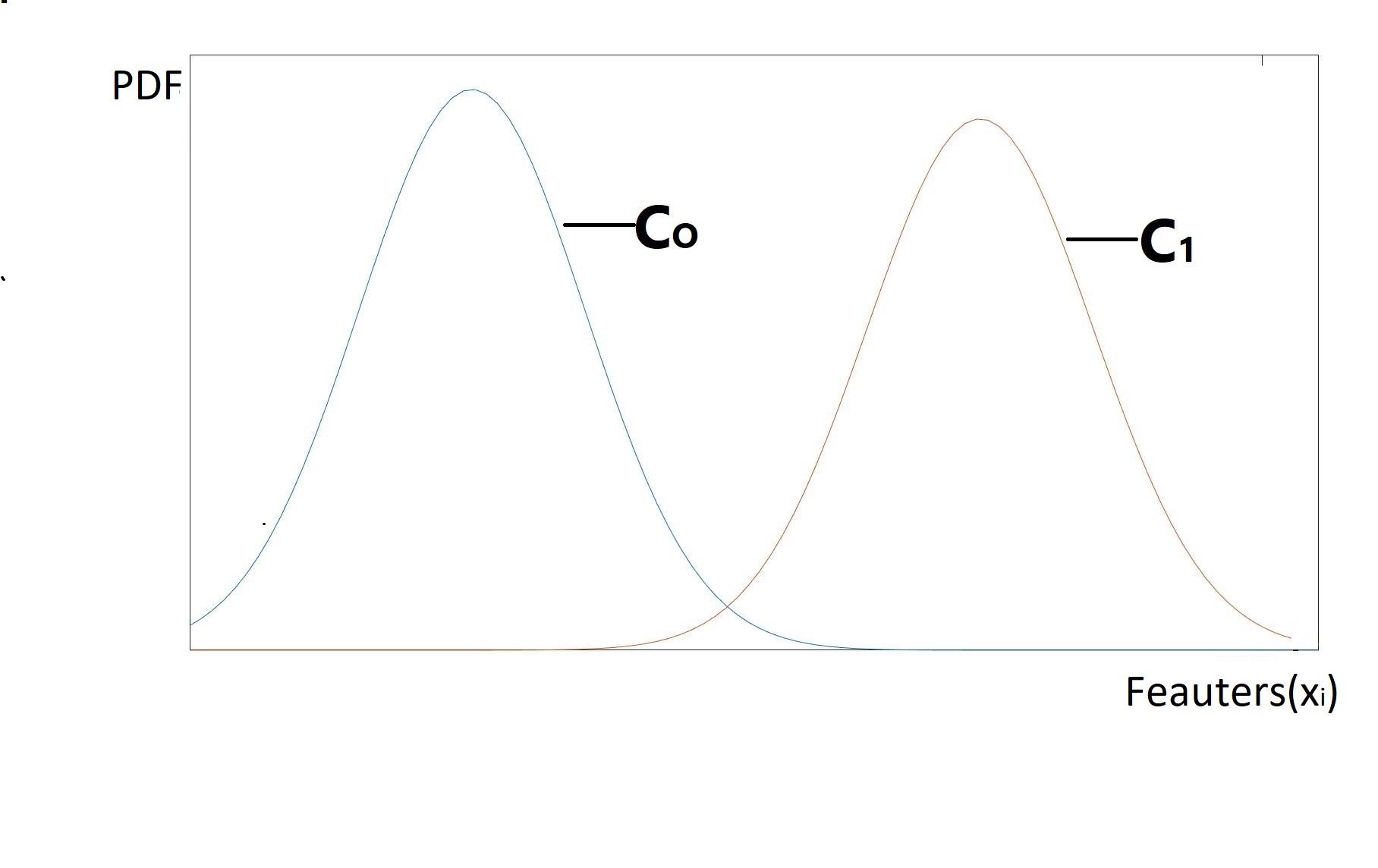}
  \caption{Two classes network}
  \label{fig:sub2}
\end{subfigure}
\caption{Features vs Probabilities densities. The x-axis in the graphs present the network features, denoted as i. The Y axis is the features probability density denoted as PDF.} 
\label{fig:test}
\end{figure}

The graph in figure 2.a illustrates the features probabilities densities of identifying $C_0$ and $C_1$ in a network trained to detect ten different classes and negative images. The active features are nearly a quarter of the total features in the network.  The area of miss-classified features is significant compared to the total areas of the features of classes $C_0$ and $C_1$ which indicates a large classification error. This is because there are many classes and the number of features dedicated to each class is small. Additional cause is there are many classes, the total number of features exceeds the upper bound number of filters optimal for this network resulting false defections. The graph in figure 2.b illustrates a network trained to detect only two classes and negative images. Most of the features detected by this network are of classes $C_0$ and $C_1$. The miss-classified feature area is small compared to the total area of both classes, indicating that the classification error is small. The number of features for each class is large enabling the training of features for detecting even more detailed features, further reducing the classification error.
In the first stage of the modular network that trained to detect general classes $C_0$ and $C_1$ ore both include in $C_g$,  a general class,  $C_g= C_0 \cup C_1$.  This eliminates the error of miss-classification between the two classes result in low classification error. Classification errors in this network are between general classes.
  
\section{Experiments}
\subsection {Implementation }

The original data set for training contains 522 original images, expanded to 46,044 training images by mirror symmetry, sharpness, brightness, and contrast augmentations and used as the training data by both the modular and the multi-class networks. The images distributed similarly between 10 classes or five pairs of similar classes: Pekinese, Spaniel, Kayak, canoe, swan, duck, sport bike, mountain bike ,Mars, Saturn and negative images with no labels. The test set contained additional 125 original images, expanded by cross validation to 647 original test images. The size of the network input images is 800x800 pixels. The multi-class network is Faster R-CNN with a backbone classification network, VGG 16, initialized by transfer learning training on ImageNet 2012 database. The building blocks of the modular network is  Faster R-CNN too with VGG 16 backbone and same initialization . Faster R-CNN networks trained on 40-50 original images for each class for various object detection tasks are prevalent \citep{fractures,cntk}. To compare between the multi-class network and the modular network,  both networks have the same hyper-parameter values previously optimized on classes other than those the networks are designated  to detect. The modular network and the multi-class network had Fine tuning training on all the networks layers. Each of the networks trained for 40 epochs, with learning rates of 0.001 on the first 10 epochs; 0.0001 on the next 10 epochs; and 0.00001 on the last 20 epochs. Both the modular and the multi-class networks inferred on this test data. Most of the original images for the training and the test sets were taken from the Caltech 101 image database and the rest were randomly chosen from the internet. 

\subsection {Experiments results }
The Faster-RCNN multi-class object detection network was trained on the data set of 46,044 images with the ten classes and the negative images. The training loss  \citep{DBLP:journals/corr/RenHG015}  is 0.0229. The multiclass network inference results are 0.87 mAP and 12\% classification error. 

The modular network has two stages. The first stage network is trained on the same data set as the multi-class network including the negative images, but the objects in the images labeled with five general classes instead of the more detailed 10 classes of the multi-class network. The modular network’s first stage classes are dog, planet, bike, boat, and bird. Each of these classes is a unification of a couple of similar classes from the 10 classes labeled for training by the multi-class network. The training loss of the modular network’s first stage is 0.0216. In the second stage, each network is trained on two fine-grained pair of classes of the multi-class network and negative images. For example, one network trains on two dog species classes, Pekinese and Spaniel and negative images, with a training loss of 0.0151 loss, while a second network is trained to detect two planets, Mars and Saturn and negative images, with a training loss of 0.0170. The modular network v1 inference results are 0.94 mAP and 4.5\% classification error. The modular network v2 inference results are 0.95 mAP and 2.5\%  classification error. These  experimental results indicate that the modular network is significantly more accurate than the multi-class network.  
\begin{table}[h!]
\centering
\caption{Object detection average precision}
\vspace{\baselineskip}
\begin{tabular}{p{4.5cm} cc}
\toprule   
\multicolumn{1}{l}{Network}&\multicolumn{1}{c}{mAP}\\
\midrule
 
Modular net v1& 0.94\\ 
Modular net v2& 0.95\\
Multi-class net (Faster R-CNN)& 0.87\\ 
First stage of the Modular net, 
general classes&0.93 \\ 
 \bottomrule
\end{tabular}
 
\label{map}
\end{table}
 Table \ref{map}   shows  experiments results of the mean average precision, mAP, of the modular networks and the multi class network, which is a regular Faster R-CNN network, tested on the same images. 
The modular network v1 mAP is calculated by taking into account the images detected as false negatives in the first stage of the modular network and thereby not appearing on the mAP of the second stage. Each false negative precision is rated as zero, and its counterpart in the calculation of the whole modular network mAP is one divided by the total number of inference images in the modular network.  In table.1 the total modular networks mAP is higher than the modular network first stage mAP even that their classification error is larger  than the first stage classification error. This means the second stage fine grained object detection networks has higher accuracy in detecting object location compared to the first stage object detection network.

Table \ref{percent} shows the experimental results of the network classification errors. The modular network classification error is significantly reduced   to 4.5\% and 2.5\%  for Modular net v1 and Modular net v2 respectively  compared to 12\%, respectively, in the multi-class network the regular Faster R-CNN . 

\begin{table}[h!] 
\centering
\caption{Classification Error}
\vspace{\baselineskip}
\begin{tabular}{p{4.5cm} cc}
\toprule
Network &  Classification error\\
\midrule
Modular net v1& 4.5\%   \\
Modular net v2&  2.5\%  \\
Multi-class net (Faster R-CNN) &  12\%  \\
First stage of the Modular net, 
general classes  & 2.25\% \\ 

\bottomrule
\end{tabular}
 
\label{percent}
\end{table} 
\vspace{\baselineskip}
\begin{figure}[h!]
     \centering
     \includegraphics[width=0.32\textwidth]{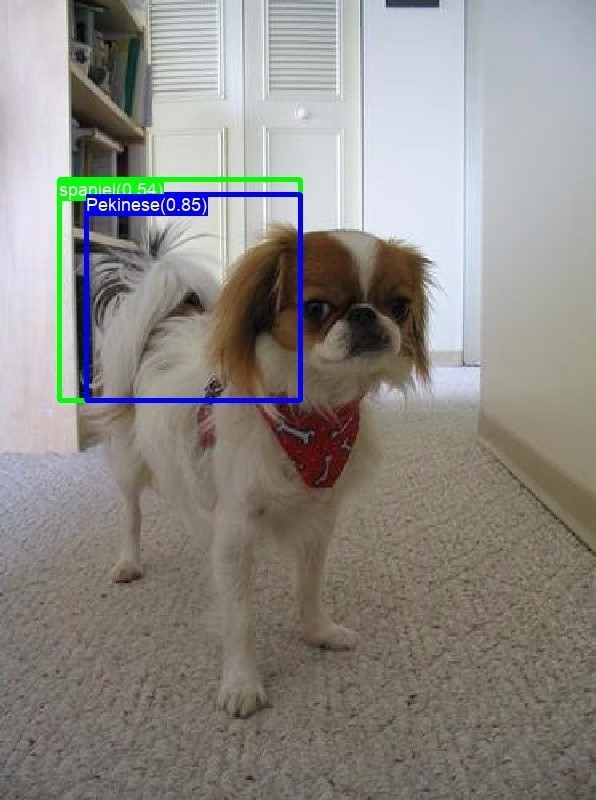}
     \includegraphics[width=0.32\textwidth]{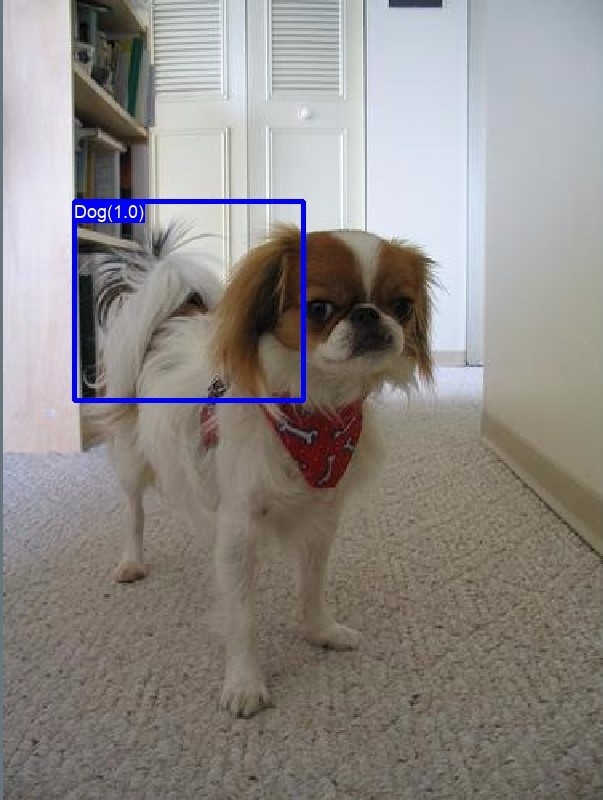}
     \includegraphics[width=0.32\textwidth]{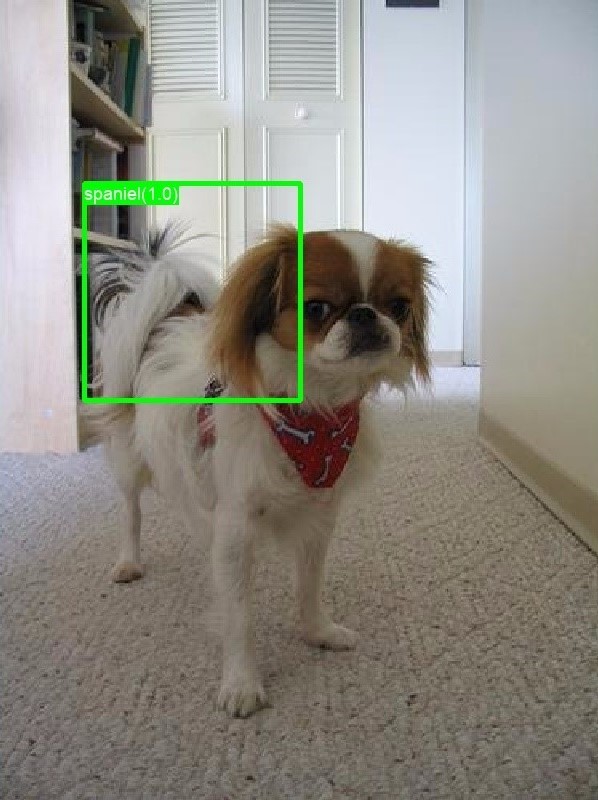}\\
     \includegraphics[width=0.32\textwidth]{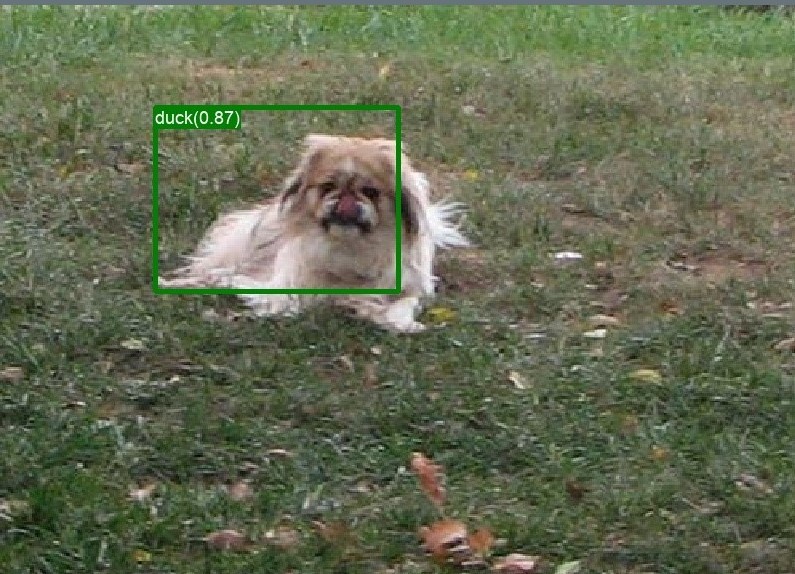}
     \includegraphics[width=0.32\textwidth]{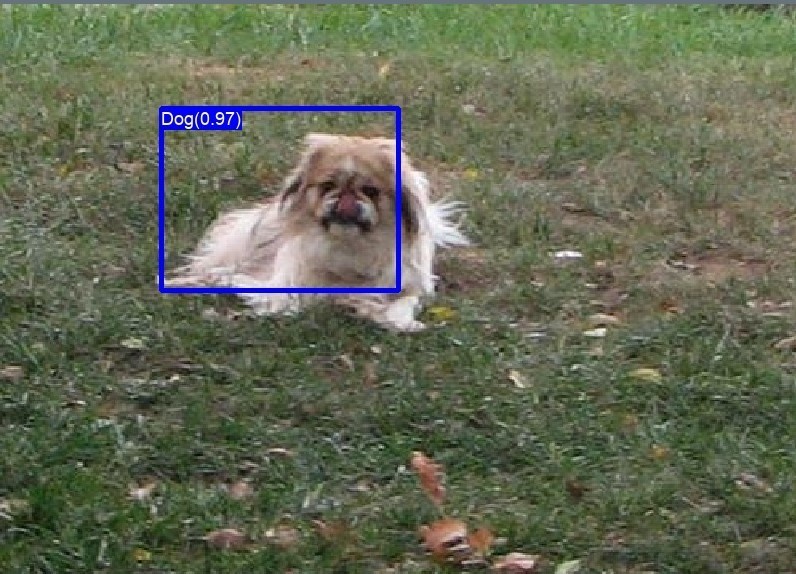}
     \includegraphics[width=0.32\textwidth]{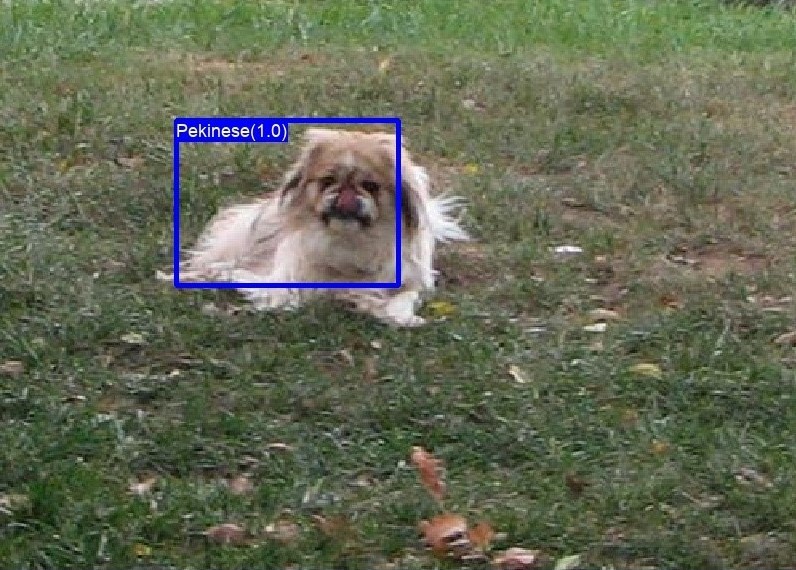}\\
     \includegraphics[width=0.32\textwidth]{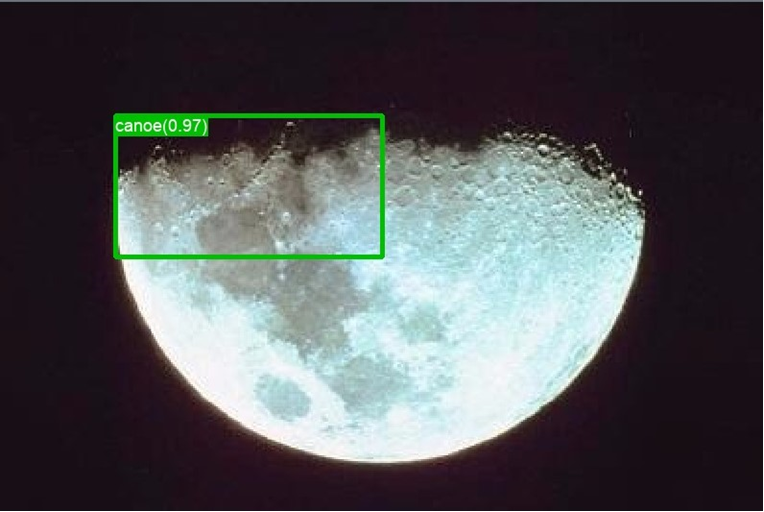}
     \includegraphics[width=0.32\textwidth]{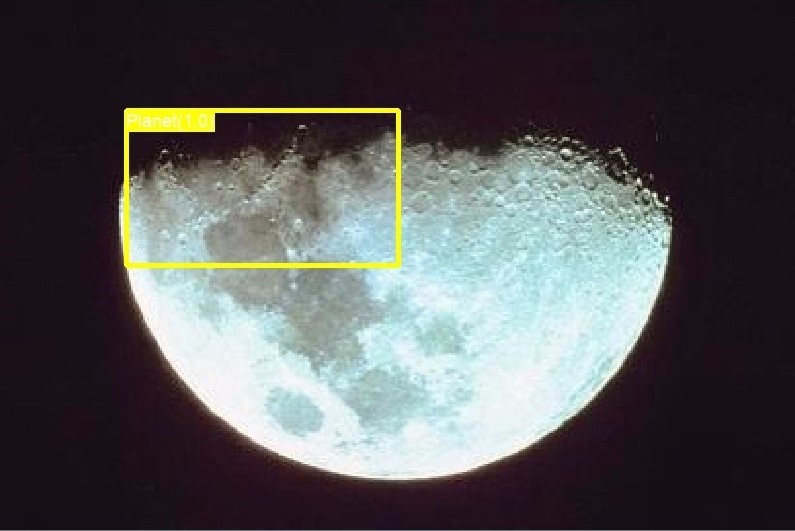}
     \includegraphics[width=0.32\textwidth]{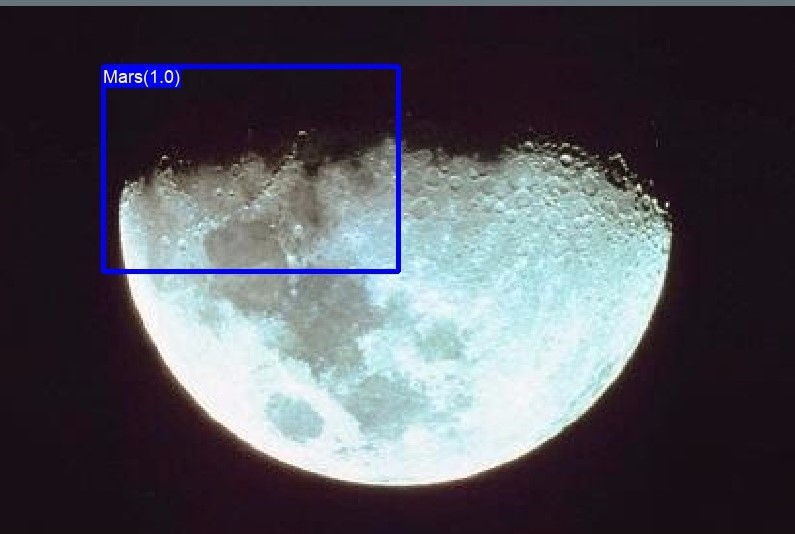}\\
     \includegraphics[width=0.32\textwidth]{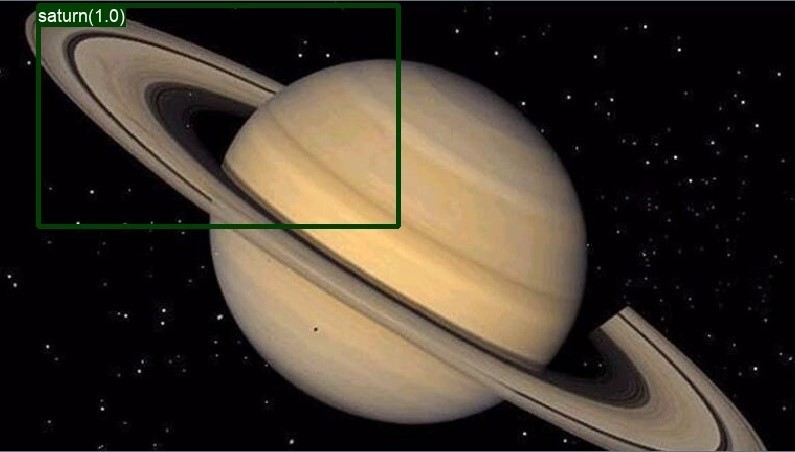}
     \includegraphics[width=0.32\textwidth]{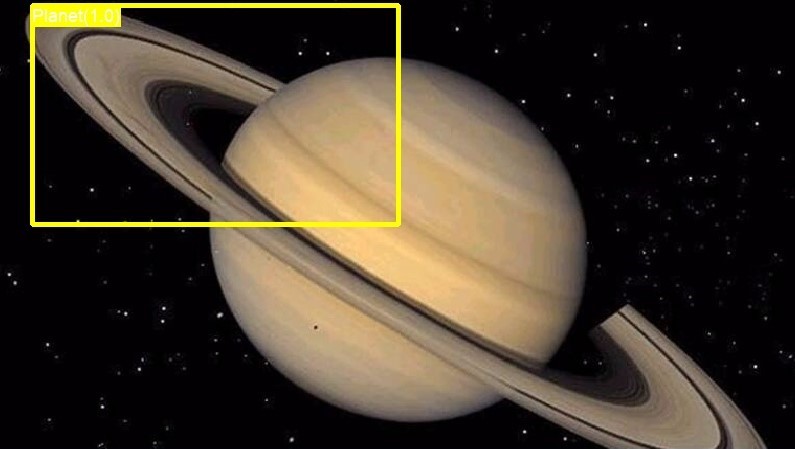}
     \includegraphics[width=0.32\textwidth]{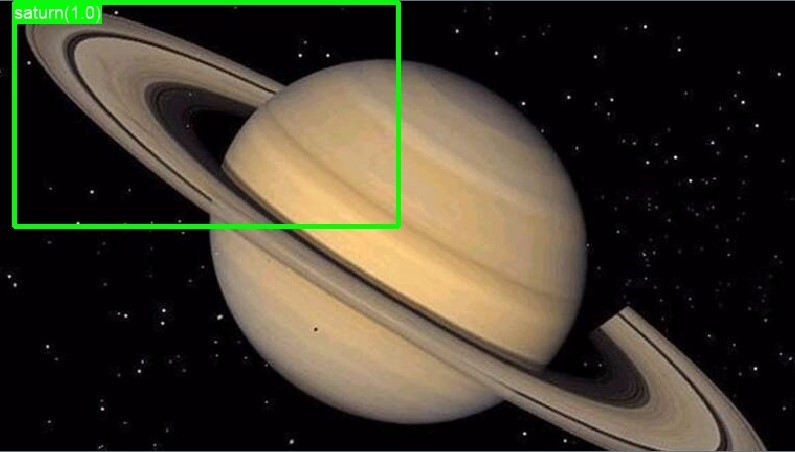}
     \caption{Left column are object detection images by the multi class network, center column are detected images by the  general classes network and right column are images detected by fine grained networks.}
     \label{fig:my_label}
 \end{figure}

In the first three rows of the first column of Figure 3, the images detected by the multi-class network have errors in classification. However, the general class network and the fine-grained network both detected the same objects correctly.
The second row of images shows that the detection of object location is more accurate in the image detected by the fine-grained network (right) compared to the image detected by the multi-class network (left).

\section{Discussion}
Our experimental result shows that network with fewer classes is more accurate. The results show that most of the classification errors in the multi-class network are between similar classes. The accuracy of the modular network for both v.1 and v.2 is higher by 7.5\%and 9.5\%, respectively, compared to the multi-class network or regular object detection network. This is a reduction of classification error by  2.7 and 4.8 times, respectively. We found the accuracy of a network that trained to detect only two similar objects is 9.5\% higher compared to the multi-class network that detects 10 classes The training results indicate that the training loss becomes smaller as the number of classes trained to be detected by the network becomes smaller. 

A fundamental question in machine learning is what kind of learning has higher accuracy?   A network trained to detect only few focused classes or one that is trained to detect many classes of a wide range of subjects. We obtained that a network initially trained on a wide range of classes by transfer learning and later trained to detect fewer classes by fine tuning on all the network layers is more accurate than a network initialized by transfer learning and later trained to detect larger numbers of classes. Previous works on transfer learning \citep{DBLP:journals/corr/HuhAE16,DBLP:journals/corr/YosinskiCBL14} determined that a network  initially trained by transfer learning and later trained to detect designated classes is more accurate compared to only being trained to detect the same designated classes. From both findings we obtain: a network initially trained by transfer learning and then designated to detect a small number of classes is more accurate compared to being trained to detect a larger number of classes with or without transfer learning.

\section{Conclusion}
The modular network presented in this paper significantly improves object detection performances in both classification and location. This is true especially for detection that requires differentiating between similar classes. This modular network improves state-of-the-art deep learning object detection networks without requiring changes in networks architecture or even hyper-parameters,  adjusting the hyper-parameters may give even higher performances. We found that reducing the number of classes a convolutional neural network is trained to detect increases the network accuracy. This modular network could be a platform for other types of deep learning networks, for example, segmentation networks, by improving their accuracy by implementing them as building blocks of the modular network. 

\bibliographystyle{plainnat} 
\bibliography{neurips_2020}

\begin{thebibliography}{33}
\providecommand{\natexlab}[1]{#1}
\providecommand{\url}[1]{\texttt{#1}}
\expandafter\ifx\csname urlstyle\endcsname\relax
  \providecommand{\doi}[1]{doi: #1}\else
  \providecommand{\doi}{doi: \begingroup \urlstyle{rm}\Url}\fi

\bibitem[B.~Juang and Lee(1997)]{568732}
W.~Hou B.~Juang and C.~Lee.
\newblock Minimum classification error rate methods for speech recognition.
\newblock \emph{IEEE Transactions on Speech and Audio Processing}, 5, 1997.

\bibitem[B.Hariharan and J.Malik(2015)]{Hariharan_2015_CVPR}
R.B~Girshick B.Hariharan, P.~Arbelaez and J.Malik.
\newblock Hypercolumns for object segmentation and fine-grained localization.
\newblock 2015.

\bibitem[Deng et~al.(2009)Deng, Dong, Socher, Li, Li, and
  Li]{Deng2009ImageNetAL}
Jia Deng, Wei Dong, Richard Socher, Li-Jia Li, Kai Li, and Fei-Fei Li.
\newblock Imagenet: A large-scale hierarchical image database.
\newblock \emph{2009 IEEE Conference on Computer Vision and Pattern
  Recognition}, pages 248--255, 2009.

\bibitem[Fukushima(1988)]{DBLP:journals/nn/Fukushima88}
K.~Fukushima.
\newblock Neocognitron: {A} hierarchical neural network capable of visual
  pattern recognition.
\newblock \emph{Neural Networks}, 1, 1988.

\bibitem[Garg and Kotecha(2018)]{inbook}
Dweepna Garg and Ketan Kotecha.
\newblock \emph{Object Detection from Video Sequences Using Deep Learning: An
  Overview}, pages 137--148.
\newblock 01 2018.
\newblock ISBN 978-981-10-4602-5.
\newblock \doi{10.1007/978-981-10-4603-2_14}.

\bibitem[He et~al.(2016)He, Zhang, Ren, and Sun]{DBLP:journals/corr/HeZRS15}
Kaiming He, Xiangyu Zhang, Shaoqing Ren, and Jian Sun.
\newblock Deep residual learning for image recognition.
\newblock In \emph{Proceedings of the IEEE conference on computer vision and
  pattern recognition}, pages 770--778, 2016.

\bibitem[Howard et~al.(2017)Howard, Zhu, Chen, Kalenichenko, Wang, Weyand,
  Andreetto, and Adam]{DBLP:journals/corr/HowardZCKWWAA17}
Andrew~G Howard, Menglong Zhu, Bo~Chen, Dmitry Kalenichenko, Weijun Wang,
  Tobias Weyand, Marco Andreetto, and Hartwig Adam.
\newblock Mobilenets: Efficient convolutional neural networks for mobile vision
  applications.
\newblock \emph{arXiv preprint arXiv:1704.04861}, 2017.

\bibitem[Huh et~al.(2016)Huh, Agrawal, and Efros]{DBLP:journals/corr/HuhAE16}
Minyoung Huh, Pulkit Agrawal, and Alexei~A Efros.
\newblock What makes imagenet good for transfer learning?
\newblock \emph{arXiv preprint arXiv:1608.08614}, 2016.

\bibitem[J.Zhang(2007)]{biobayes}
H.~Deng J.Zhang.
\newblock Gene selection for classification of microarray data based on the
  bayes error.
\newblock \emph{BMC Bioinformatics}, 8, 2007.

\bibitem[K.~Jonathan and FLi(2015)]{Krause_2015_CVPR}
Y.~Jianchao K.~Jonathan, J.~Hailin and FLi.
\newblock Fine-grained recognition without part annotations.
\newblock In \emph{The IEEE Conference on Computer Vision and Pattern
  Recognition (CVPR)}, June 2015.

\bibitem[karol zak(2018)]{cntk}
karol zak.
\newblock Cntk-hotel-pictures-classificator.
\newblock \emph{https://github.com/karolzak/cntk-hotel-pictures-classificator},
  2018.

\bibitem[Karpathy(2014)]{Karpathy_2014_CVPR}
A.~Karpathy.
\newblock Large-scale video classification with convolutional neural networks.
\newblock In \emph{The IEEE Conference on Computer Vision and Pattern
  Recognition (CVPR)}, June 2014.

\bibitem[Krizhevsky et~al.(2012)Krizhevsky, Sutskever, and
  Hinton]{Krizhevsky2012ImageNetCW}
Alex Krizhevsky, Ilya Sutskever, and Geoffrey~E. Hinton.
\newblock Imagenet classification with deep convolutional neural networks.
\newblock In \emph{NIPS}, 2012.

\bibitem[Käding et~al.(2017)Käding, Rodner, Freytag, and Denzler]{inproceed}
Christoph Käding, Erik Rodner, Alexander Freytag, and Joachim Denzler.
\newblock Fine-tuning deep neural networks in continuous learning scenarios.
\newblock pages 588--605, 03 2017.
\newblock \doi{10.1007/978-3-319-54526-4_43}.

\bibitem[Lecler et~al.(2019)Lecler, Duron, Balvay, Savatovsky, Bergès, Zmuda,
  Farah, Galatoire, Bouchouicha, and Fournier]{articl}
A.~Lecler, L.~Duron, Daniel Balvay, Julien Savatovsky, Olivier Bergès, Mathieu
  Zmuda, Edgard Farah, O.~Galatoire, A.~Bouchouicha, and Laure Fournier.
\newblock Combining multiple magnetic resonance imaging sequences provides
  independent reproducible radiomics features.
\newblock \emph{Scientific Reports}, 9, 02 2019.
\newblock \doi{10.1038/s41598-018-37984-8}.

\bibitem[Lin et~al.(2014)Lin, Maire, Belongie, Hays, Perona, Ramanan,
  Doll{\'a}r, and Zitnick]{lin2014microsoft}
Tsung-Yi Lin, Michael Maire, Serge Belongie, James Hays, Pietro Perona, Deva
  Ramanan, Piotr Doll{\'a}r, and C~Lawrence Zitnick.
\newblock Microsoft coco: Common objects in context.
\newblock In \emph{European conference on computer vision}, pages 740--755.
  Springer, 2014.

\bibitem[Liu et~al.(2016)Liu, Anguelov, Erhan, Szegedy, Reed, Fu, and
  Berg]{journals/corr/LiuAESR15}
Wei Liu, Dragomir Anguelov, Dumitru Erhan, Christian Szegedy, Scott Reed,
  Cheng-Yang Fu, and Alexander~C. Berg.
\newblock Ssd: Single shot multibox detector.
\newblock pages 21--37, 2016.

\bibitem[Oquab et~al.(2014)Oquab, Bottou, Laptev, and Sivic]{oquab2014learning}
Maxime Oquab, Leon Bottou, Ivan Laptev, and Josef Sivic.
\newblock Learning and transferring mid-level image representations using
  convolutional neural networks.
\newblock In \emph{Proceedings of the IEEE conference on computer vision and
  pattern recognition}, pages 1717--1724, 2014.

\bibitem[Pan and Yang(2009)]{IEEE_Transactions}
Sinno~Jialin Pan and Qiang Yang.
\newblock Ieee transactions on knowledge and data engineering.
\newblock \emph{CoRR}, 22, 2009.

\bibitem[Qiu et~al.(2019)Qiu, Huang, and Lee]{qiu2019visual}
Jielin Qiu, Ge~Huang, and Tai~Sing Lee.
\newblock Visual sequence learning in hierarchical prediction networks and
  primate visual cortex.
\newblock In \emph{Advances in Neural Information Processing Systems}, pages
  2658--2669, 2019.

\bibitem[Redmon et~al.(2016)Redmon, Divvala, Girshick, and
  Farhadi]{DBLP:journals/corr/RedmonDGF15}
Joseph Redmon, Santosh Divvala, Ross Girshick, and Ali Farhadi.
\newblock You only look once: Unified, real-time object detection.
\newblock In \emph{Proceedings of the IEEE conference on computer vision and
  pattern recognition}, pages 779--788, 2016.

\bibitem[{Ren} et~al.(2017){Ren}, {He}, {Girshick}, and
  {Sun}]{DBLP:journals/corr/RenHG015}
S.~{Ren}, K.~{He}, R.~{Girshick}, and J.~{Sun}.
\newblock Faster r-cnn: Towards real-time object detection with region proposal
  networks.
\newblock \emph{IEEE Transactions on Pattern Analysis and Machine
  Intelligence}, 39\penalty0 (6):\penalty0 1137--1149, June 2017.
\newblock ISSN 1939-3539.
\newblock \doi{10.1109/TPAMI.2016.2577031}.

\bibitem[Ross(2015)]{journals/corr/Girshick15}
Girshick Ross.
\newblock Fast r-cnn.
\newblock In \emph{Proceedings of the IEEE international conference on computer
  vision}, pages 1440--1448, 2015.

\bibitem[Russakovsky et~al.(2015)Russakovsky, Deng, Su, Krause, Satheesh, Ma,
  Huang, Karpathy, Khosla, Bernstein, Berg, and
  Fei-Fei]{10.1007/s11263-015-0816-y}
Olga Russakovsky, Jia Deng, Hao Su, Jonathan Krause, Sanjeev Satheesh, Sean Ma,
  Zhiheng Huang, Andrej Karpathy, Aditya Khosla, Michael Bernstein,
  Alexander~C. Berg, and Li~Fei-Fei.
\newblock Imagenet large scale visual recognition challenge.
\newblock \emph{Int. J. Comput. Vision}, 115\penalty0 (3), 2015.

\bibitem[{Salakhutdinov} et~al.(2011){Salakhutdinov}, {Torralba}, and
  {Tenenbaum}]{5995720}
R.~{Salakhutdinov}, A.~{Torralba}, and J.~{Tenenbaum}.
\newblock Learning to share visual appearance for multiclass object detection.
\newblock In \emph{CVPR 2011}, pages 1481--1488, 2011.

\bibitem[Simonyan and Zisserman(2014)]{simonyan2014very}
K.~Simonyan and A.~Zisserman.
\newblock Very deep convolutional networks for large-scale image recognition.
\newblock \emph{arXiv preprint arXiv:1409.1556}, 2014.

\bibitem[Singh et~al.(2016)Singh, Marks, Jones, Tuzel, and
  Shao]{Singh_2016_CVPR}
Bharat Singh, Tim~K Marks, Michael Jones, Oncel Tuzel, and Ming Shao.
\newblock A multi-stream bi-directional recurrent neural network for
  fine-grained action detection.
\newblock In \emph{Proceedings of the IEEE Conference on Computer Vision and
  Pattern Recognition}, pages 1961--1970, 2016.

\bibitem[Szegedy et~al.(2015)Szegedy, Liu, Jia, Sermanet, Reed, Anguelov,
  Erhan, Vanhoucke, and Rabinovich]{43022}
Christian Szegedy, Wei Liu, Yangqing Jia, Pierre Sermanet, Scott Reed, Dragomir
  Anguelov, Dumitru Erhan, Vincent Vanhoucke, and Andrew Rabinovich.
\newblock Going deeper with convolutions.
\newblock In \emph{Proceedings of the IEEE conference on computer vision and
  pattern recognition}, pages 1--9, 2015.

\bibitem[Xiaomin and Yaming(2006)]{article2}
Bao Xiaomin and Wang Yaming.
\newblock Apple image segmentation based on the minimum error bayes decision
  [j].
\newblock \emph{Transactions of the Chinese Society of Agricultural
  Engineering}, 5, 2006.

\bibitem[Yahalomi(2019)]{rplan}
Erez Yahalomi.
\newblock Deep learning networks for medical images.
\newblock \emph{PhD Research plan}, March 2019.

\bibitem[Yahalomi et~al.(2019)Yahalomi, Chernofsky, and Werman]{fractures}
Erez Yahalomi, Michael Chernofsky, and Michael Werman.
\newblock Detection of distal radius fractures trained by a small set of x-ray
  images and faster r-cnn.
\newblock In \emph{Intelligent Computing. CompCom 2019. Advances in Intelligent
  Systems and Computing}, volume 997, pages 971--981, 2019.

\bibitem[Yang and Hu(2012)]{article1}
Shuang~Hong Yang and Bao-Gang Hu.
\newblock Discriminative feature selection by nonparametric bayes error
  minimization.
\newblock \emph{IEEE Transactions on knowledge and data engineering},
  24\penalty0 (8):\penalty0 1422--1434, 2012.

\bibitem[Yosinski et~al.(2014)Yosinski, Clune, Bengio, and
  Lipson]{DBLP:journals/corr/YosinskiCBL14}
Jason Yosinski, Jeff Clune, Yoshua Bengio, and Hod Lipson.
\newblock How transferable are features in deep neural networks?
\newblock In \emph{Advances in neural information processing systems}, pages
  3320--3328, 2014.

\end{thebibliography}

\end{document}